\documentclass[letterpaper]{article} 
\usepackage{aaai23}  
\usepackage{times}  
\usepackage{helvet}  
\usepackage{courier}  
\usepackage[hyphens]{url}  
\usepackage{graphicx} 
\usepackage{natbib}  
\usepackage{caption} 
\usepackage{algorithm}
\usepackage{algorithmic}
\usepackage{booktabs}
\usepackage{multirow}

\usepackage{newfloat}
\usepackage{listings}
\DeclareCaptionStyle{ruled}{labelfont=normalfont,labelsep=colon,strut=off} 
\floatstyle{ruled}
\newfloat{listing}{tb}{lst}{}
\floatname{listing}{Listing}
\title{VideoDubber: Machine Translation with Speech-Aware Length Control for Video Dubbing}
\author{
    Yihan Wu\textsuperscript{\rm 1}\thanks{Supported by the Outstanding Innovative Talents Cultivation Funded Programs 2023 of Renmin University of China.}, Junliang Guo\textsuperscript{\rm 2}, Xu Tan\textsuperscript{\rm 2},
    Chen Zhang\textsuperscript{\rm 3},
    Bohan Li\textsuperscript{\rm 3},
    Ruihua Song\textsuperscript{\rm 1}\footnote{Corresponding author},\\ Lei He\textsuperscript{\rm 3}, Sheng Zhao\textsuperscript{\rm 3}, Arul Menezes\textsuperscript{\rm 4}, Jiang Bian\textsuperscript{\rm 2}
}
\affiliations{
    \textsuperscript{\rm 1} Gaoling School of Artificial Intelligence, Renmin University of China\\
    \textsuperscript{\rm 2} Microsoft Research Asia,\\
    \textsuperscript{\rm 3} Microsoft Azure Speech,\\ \textsuperscript{\rm 4} Microsoft Azure Translation \\
    yihanwu@ruc.edu.cn, \{junliangguo, xuta, zhangche, bohli, helei, szhao, arulm, jiang.bian\}@microsoft.com
}

\usepackage{bibentry}

\begin{document}

\maketitle
\begin{abstract}

Video dubbing aims to translate the original speech in a film or television program into the speech in a target language, which can be achieved with a cascaded system consisting of speech recognition, machine translation and speech synthesis.
To ensure the translated speech 
to be 
well aligned with the corresponding video, the length/duration of the translated speech should be as close as possible to that of the original speech, which requires strict length control.
Previous works usually control the number of words or characters generated by the machine translation model to be similar to the source sentence, without considering the isochronicity of speech as 
the speech duration of words/characters in different languages varies.
In this paper, we propose VideoDubber, a machine translation system tailored for the task of video dubbing, which directly considers the
speech duration of each token in translation, to match the length of source and target speech.
Specifically, we 
control the speech length of generated sentence by guiding the prediction of each word with the duration information, including the speech duration of itself as well as how much duration is left for the remaining words.
We design experiments on four language directions (German $\rightarrow$ English, Spanish $\rightarrow$ English, Chinese $\leftrightarrow$ English), and the results show that VideoDubber achieves better length control ability on the generated speech than baseline methods.
To make up the lack of real-world datasets, we also construct a real-world test set collected from films to provide comprehensive evaluations on the video dubbing task. 

\end{abstract}

\section{Introduction}

Video dubbing\footnote{There are usually two scenarios for video dubbing: 1) translating the speech of a video from one language to
another; 2) generating speech in the same language according to the text transcripts of a video. In this paper, we focus on the first video dubbing for translation scenario.} aims to translate the original speech in a movie or TV from a source language to the speech in a target language.
Generally, automatic video dubbing system
consists of
three cascaded sub-tasks, i.e., Automatic Speech Recognition~(ASR)~\cite{yu2016automatic}, Neural Machine Translation~(NMT)~\cite{DBLP:conf/nips/VaswaniSPUJGKP17} and Text-to-Speech~(TTS)~\cite{tan2021survey,tan2022naturalspeech}. Specifically, ASR transcribes the original speech into text in the source language, which can be omitted when subtitles are available. NMT aims to translate text in the source language to the target language, and TTS finally synthesizes
the translated text into target speech.
Different from speech-to-speech translation~\cite{wahlster2013verbmobil}, video dubbing requires strict isochronous constraints between the original source speech and the synthesized target speech to ensure that the speech matches the original video footage in terms of length and thus provide immersive watching feeling, which brings additional challenges.

In the cascaded video dubbing system, both NMT and TTS can determine the length of translated speech, i.e., the number of translated words is determined by the NMT model, while the length of the synthesized speech can be controlled by TTS via adjusting the pause and duration of words. 
Previous automatic video dubbing systems control the speech isochrony separately,
by first controlling the number of generated target words/characters similar to that of the source words, and then adjusting the speaking rate to ensure the duration of synthesized speech similar to that of the source speech. 
However, as the speech duration of words/characters in different languages varies, the same number of words/characters in the source and target sentences does not guarantee the same length of speech. Therefore, TTS has to adjust the speaking rate of each word in a wide range to match the total speech length, which will affect the fluency and naturalness of synthesized speech (e.g., the speaking rates may be inconsistent with those of adjacent sentences). 

\begin{figure*}[t]
\centering
\includegraphics[width=0.8\linewidth]{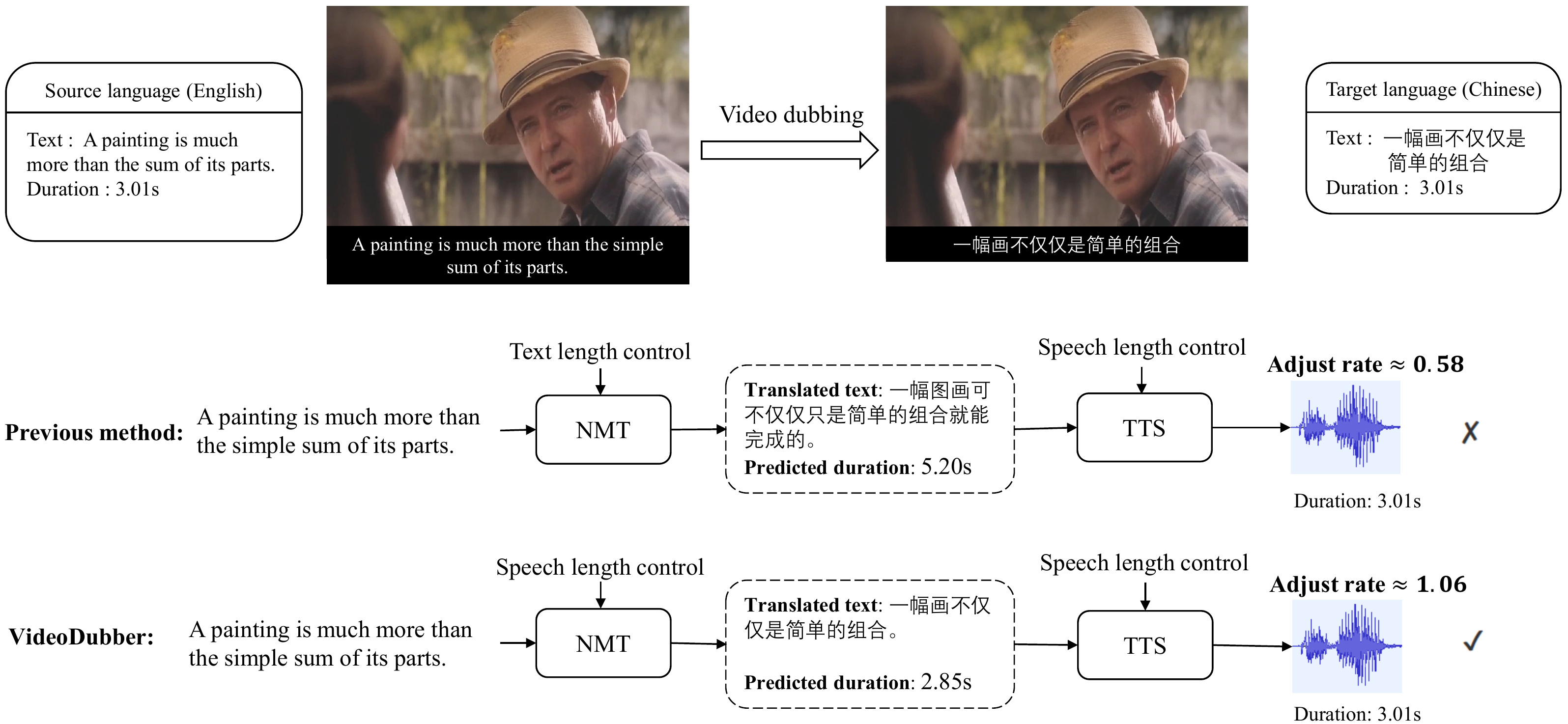} 
\caption{An example of video dubbing from English to Chinese. When translating the source sentence ``A painting is much more than the simple sum of its parts.'' in NMT stage, its corresponding speech duration is 3.01s in original speech. Previous method control the number of words of translated sentence, but the length of corresponding speech is very different from the original speech without adjustment. The translation result from our method considering speech length directly, thus generate translated speech that are very close to the original one. Then the TTS model generates the desired translation result to speech with a little duration adjustment. }.
\label{fig1}
\end{figure*}

In this paper, we tackle the length control problem in a 
systematic way. 
Our idea starts from a simple motivation: matching the length of the synthesized speech with the original one while avoiding speaking rate inconsistency of adjacent sentences.
To preserve high hearing quality, 
we should assign the responsibility for
length control more to NMT and less to TTS, which requires NMT can be controlled by original speech length while maintaining translation quality.
To achieve this goal, we propose the following techniques. 
Firstly, 
to control the speech duration when translating the target sentence, we make the model aware of how much duration is left at each time-step as well as the speech duration of each word, by representing it as 
two kinds of positional embeddings which are added to the original one and taken as input of the decoder. 
Secondly, 
we introduce a special pause word \texttt{[P]} which is inserted between each target word, in order to 
control the speech length more smoothly by  
considering
the prosody (language-related tempo, rhythm and pause) of speech through adjusting the duration of \texttt{[P]}.
Finally, while training, we calculate the alignment between the target speech and text to obtain the speech duration of each target word instead of using only the number of words/characters. Also 
we introduce a neural network component named duration predictor to explicitly model the speech duration of each word. In inference, given the source speech length, the duration predictor works together with the decoder to generate high quality translations with similar speech length to the source.
Moreover, considering the scarcity of real-world video dubbing dataset~(i.e., motion pictures with golden cross-lingual source and target speech available), we construct a test set collected from dubbed films to provide comprehensive evaluations of video dubbing systems.

The main contributions of this work are summarized as follows:
\begin{itemize}
    \item We propose VideoDubber, a machine translation system tailored for the task of video dubbing, which directly controls the speech length of generated sentence by calculating and encoding the speech duration of each word into the model. 
    \item We conduct experiments on four language directions, with both objective and subjective evaluation metrics. Experiment results shows that VideoDubber achieves better length control compared with baseline method in all languages.
    \item Due to the scarcity of the video dubbing test dataset, we construct a real-world test set collected from films to provide comprehensive evaluation on the video dubbing task.
\end{itemize}

\begin{figure*}[h]
\centering
\includegraphics[width=0.65\linewidth]{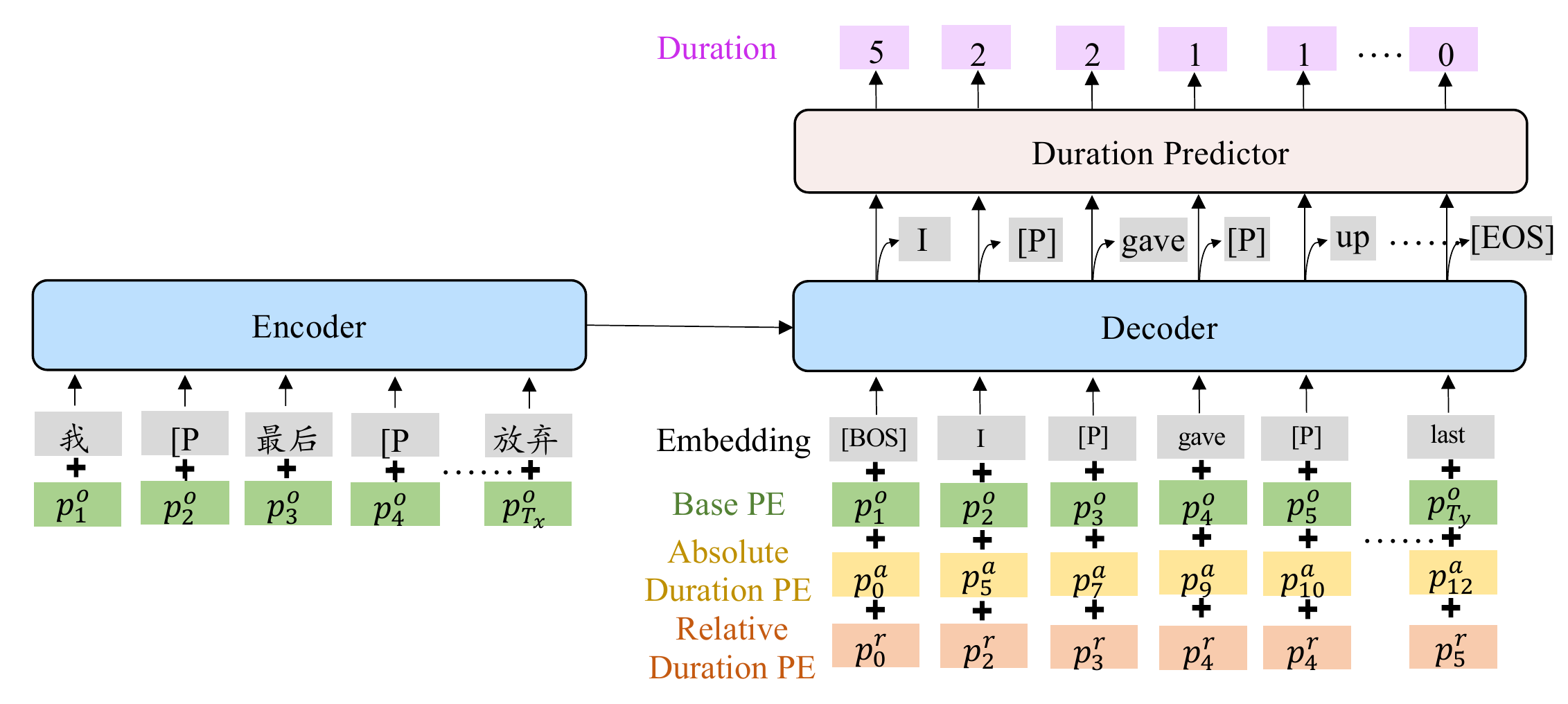} 
\caption{The overall architecture of VideoDubber with speech-aware length control for video dubbing. PE stands for positional embedding and \texttt{[P]} indicates the special pause token. We set $N=5$ and follow Equation~(\ref{equ:pos_r}) to calculate the relative duration PE.}
\label{fig2}
\end{figure*}

\section{Background}

We first provide the problem definition of the video dubbing task.
\paragraph{Problem Definition}
Given the source speech $s$ with duration $T_s$ and its transcription $x=(x_1, ..., x_{T_x})$, we aim to generate the translated text sequence $y=(y_1, ..., y_{T_y})$ as well as the synthesized speech $t$ with duration $T_t$, where $T_t$ should be as close to $T_s$ as possible, while maintaining the high translation quality of $y$ as well as the rhythmic and fluency of $t$.

Existing video dubbing works~\cite{oktem2019pa4md,federico2020evapa,lakew2021mtverb,virkar2021improvepa,sharma2021intratts,effendi2022durationtts,lakew2022isometricmt,virkar2022offscreenpa,derek2022isochronyMT} are usually based on a cascaded speech-to-speech translation system~\cite{federico2020stst2vd} with ad-hoc designs, mainly concentrating on the Neural Machine Translation~(NMT) and Text-To-Speech~(TTS) stages.
In the NMT stage, related works achieve the length control by
assuming that similar number of words/characters should have similar speech length, and therefore
encourage a model to generate target sequence with similar number of words/characters to the source sequence~\cite{federico2020stst2vd}. 
They compute the 
length ratio between source and target text sequences as the verbosity information during training, and control translated speech length by setting length ratio to $1$ during inference.
\citet{lakew2022isometricmt} introduce a character-level length constraint to the NMT model, by augmenting the training set with length controlled back translation.
Moreover, following previous works~\citep{takase2019pecontrol,lakew2019controlmt}, \citet{derek2022isochronyMT} encode the number of characters that are going to be generated into the position embedding to incorporate the length control regularization.
These methods only control the number of 
generated 
words/characters to be similar to the source sequence, without considering the discrepancy of the speech duration of words/characters in source and target languages.

The TTS stage
synthesizes the translated text sequence to speech with 
isochronous constraints, by adjusting the speaking rate to ensure the duration of the generated speech fits the source speech~\cite{sharma2021intratts,effendi2022durationtts}.
Further, \citet{effendi2022durationtts} present a duration model which can predict and adjust the phoneme duration by uniform normalization or non-isoelastic normalization~\cite{effendi2022durationtts}. To relax the duration adjustment effort and match the pauses and phrases in the target to the source speech, several works~\citep{federico2020stst2vd,federico2020evapa,virkar2021improvepa} introduce an additional
prosodic alignment module before the TTS stage. It inserts pause tokens into the translated text considering the speaking rate match between corresponding source-target phrases and the linguistic plausibility of the chosen split points. However, this two-stage rule-based approach cannot take advantage of natural pauses and rhythm from speech directly. Moreover, it only inserts pauses between phrases, ignoring that the pauses in different granularities usually correspond to different durations.
As the audience is sensitive to the speaking rate change of speech, the inconsistent speaking rates of adjacent sentences,
e.g., faster at the previous sentence while slower at the current one,
will destroy the rhythm and fluency
of the speech. Therefore, in this paper, we try to alleviate the burdens of length control on TTS and concentrate on the NMT stage, by proposing a speech-aware length control model to directly control the speech duration instead of the number of tokens of the generate target sequence.


\section{Proposed Method}
We introduce VideoDubber in this section. 
Specifically, we
control the speech length of generated sentence by guiding
the prediction of each word with duration information. An illustration of the model architecture is shown in Figure~\ref{fig2}.

\subsection{Overview}
To match the length of synthesized speech with the original one while maintaining fluency and naturalness, we should assign the responsibility for length control more to NMT and less to TTS. Thus, the goal becomes how to achieve speech-aware length control on NMT while maintaining high translation quality. To achieve this goal,
we first obtain the speech duration $d = (d_1, ..., d_{T_y})$ of each target word and then incorporate it into the NMT model.

Specifically, we integrate the duration information by designing two kinds of duration-aware positional embeddings. One is \textit{absolute} duration positional embedding which indicates the accumulated duration information at the current time-step, the other is \textit{relative} duration positional embedding calculated as the ratio between the absolute duration and the total duration, indicating how much duration is left for future tokens. In this way, the model is trained to jointly consider the semantic and duration information when making predictions.
We also take pauses into consideration to control the speech length with more flexibility by introducing a special pause token  \texttt{[P]}.

To obtain the speech duration for each target word, while training, we first calculate the alignment between speech and text in phoneme-level through Montreal forced alignment (MFA)~\cite{mcauliffe2017mfa}\footnote{MFA is an open-source tool for speech-text alignment with good performance.}.
Then, we introduce a duration predictor, which is a neural network component consisting of convolutional layers and being inserted on the top of the decoder, to predict the duration of each word conditioned on the hidden output of decoder. 
Next in inference, given a total speech length~(i.e., the length of source speech), the decoder will determine the appropriate translation conditioned on both the semantic representation and the duration information
at each step in an autoregressive manner as shown in Figure~\ref{fig2}. 
We dive into details in the following section.

\subsection{Model Architecture}
\label{sec3.2}
\subsubsection{Duration-aware Position Embedding}
To take account of the speech duration in translation, we integrate the duration information via designing duration-aware positional embeddings for each token. Here, we design two kinds of positional embedding
to provide the absolute and relative speech-aware duration information respectively.

We denote the original positional embedding in Transformer~\citep{Transformer2017} as $p^o$.
Then given the $i$-th target token $y_i$ and its duration $d_i$, 
we define the absolute duration positional embedding $p^a_i$ by the accumulated speech duration up-to-now, which can be written as:
\begin{equation}
    p^a_i = PE(\sum_{j=1}^{i} d_j),
\end{equation}
where $PE(\cdot)$ indicates the lookup function over the sinusoidal positional embedding matrix.

The introduced
$p^a$ tells the model the up-to-now speech length of generated words. 
Moreover,
to make the model aware of how much duration is left and therefore plan the generation by choosing words with the appropriate duration,
we also introduce a
relative duration positional embedding $p^r$ to incorporate the global duration information.
It is calculated as the ratio between the accumulated duration and the total duration:
\begin{equation}
\label{equ:pos_r}
    p^r_i = PE( q_{N} ( {\frac{\sum_{j=0}^{i} d_j}{\sum_{j=0}^{T_y} d_j}} )),
\end{equation}
where 
$q_{N} (x) = \lfloor x \times N \rfloor$ quantizes the float duration ratio from $[0,1] $ into integers within $[0,N]$.
Finally, given the word embedding of the $i$-th target token $w_i$, we add the three positional embeddings with it to construct the input to decoder:
\begin{equation}
    h_i = w_i + p^o_i + p^a_i + p^r_i.
\end{equation}
The proposed two position embeddings are complementary to each other, and we evaluate their effectiveness
in experiments.


\subsubsection{Pause Token}
We further consider the pauses between words and adjust their duration to ensure the flexibility of length control, i.e., we can increase or decrease the duration of pauses automatically to share some responsibility for length control, which can avoid uneven speaking rate between adjacent sentences and is crucial for the fluency of speech. As a byproduct, pauses can also be utilized to model the prosody of the speech.
Specifically, we explicitly model the pauses in speech by introducing a special token \texttt{[P]}, which is inserted between each word~(instead of sub-words) in both the source and target sentences. For example, the phrase ``\texttt{A painting}'' is tokenized as ``\texttt{A pain@@ ting}'' after conducting Byte-Pair Encoding~(BPE)~\citep{sennrich2015neural}, which is then transformed to ``\texttt{A [P] pain@@ ting}'' after inserting the pause token. 
The duration of the pause token can also be obtained through MFA, and it varies in different context.
In this way, the model learns to predict a pause token after each word with appropriate duration in context,
providing a more flexible way for speech length control.

\subsubsection{Duration Predictor}
Following FastSpeech 2~\citep{ren2021fs2}, we use MFA tool to compute ground truth word duration for training.
To make the model able to predict the speech duration of each token when the ground truth duration is absent in inference, we introduce an extra neural network component called duration predictor, which
is built on top of the decoder.
While training, it takes the hidden representation of each token as input, and 
predicts the golden duration of each token, which represents how many mel-frames correspond to this token, and is converted into the logarithmic domain for ease of prediction. 
The duration predictor is optimized with the mean square error (MSE) loss, taking the ground truth duration as training target. 
Denoting the duration sequence as $d=(d_1, ..., d_{T_y})$, the prediction can be written as:
\begin{equation}
    p(d|x,y) = \prod_{i=1}^{T_y} p(d_i|x, y_{<i};\theta^{du}),
\end{equation}
where $\theta_{du}$ indicates the parameters of duration predictor.

\subsection{Discussion}
\label{sec:discuss}

\paragraph{Overall Length Control Pipeline }
After text translation with speech-aware length control as we introduced in Section~\ref{sec3.2}, we further achieve more precise length control through duration adjustment in TTS stage. We employ AdaSpeech 4~\cite{ada4}, a zero-shot TTS model which can adjust duration directly.
Instead of uniform duration adjustment which scales the duration of each phoneme by a same factor, we only adjust the duration of vowels and keep the length of consonants unchanged, since the duration of consonants is not significantly
changed when people slow down or speed up their speech in natural~\cite{kruspe2015speechdata}. With this systematic approach on NMT and TTS, we can achieve precise and flexible speech length control, so as to ensure the translated speech to be well aligned with the corresponding video in video dubbing scenario.

\paragraph{About the Pause Token}
Some related works~\citep{federico2020stst2vd,virkar2021improvepa} also introduce the pause token into the video dubbing framework, by inserting the pause token between phrases in the translated text with a dynamic programming algorithm. VideoDubber differs from theirs from two perspectives. Firstly, we provide more fine-grained control of the pause by inserting the pause token into each word instead of phrase. Secondly, the speech duration of each pause token can be different in different context and end-to-end learned by the model, which provides better flexibility on length control and prosody modeling of synthesized speech.

\paragraph{About Video Dubbing Dataset}
Due to the scarcity of video dubbing dataset, we usually conduct experiments on speech-to-speech translation datasets, where the lengths of the source and target speech in the training and test data are not guaranteed to be similar. Thus, this is actually different from the real-world video dubbing, where the length of source and target speech exactly match. It is not accurate to evaluate the semantic translation quality of the video dubbing system using the reference translation that is not in same length of the source speech (i.e., the controlled speech length in NMT model). Therefore, we try to alleviate this issue in two aspects: 1) In evaluation, we use the length of source and target speech respectively to control the NMT model, where the length of source speech is consistent with what we practically use in video dubbing, and the length of target speech is used to evaluate how our model performs when the length of reference sentence is the same as the control length. 2) We additionally construct a real-world video dubbing test set from dubbed films, where the source and target speech have exactly the same length. In this way, the constructed test set can be used to evaluate the video dubbing system in practical scenarios. 

Another potential issue is that the NMT model is trained to generate text with the target speech length (which is different from the source speech length), but in inference, it is used to generate text with the source speech length, which causes mismatch. However, it is really hard to mitigate this issue since large-scale training data with the same source and target length is very difficult to collect. Nevertheless, we plan to construct a training dataset with the same speech length in the source and target utterances by knowledge distillation, where the distilled target sentences are controlled by our NMT model to have the same speech length with the source sentences. We leave it for future work.


\begin{table*}[h!]
\small
\centering
\begin{tabular}{l|c|c|c|c|c|c|c|c|c|c|c|c}
\toprule
\multirow{3}*{Settings} &\multicolumn{3}{c|}{Es-En} & \multicolumn{3}{c|}{De-En} & \multicolumn{3}{c|}{Zh-En} &  \multicolumn{3}{c}{En-Zh} \\
\cmidrule{2-13}
 & \multirow{2}*{BLEU} & \multicolumn{2}{c|}{$\textrm{SLC}_{p}$} & \multirow{2}*{BLEU} & \multicolumn{2}{c|}{$\textrm{SLC}_{p}$} & \multirow{2}*{BLEU} & \multicolumn{2}{c|}{$\textrm{SLC}_{p}$}  & \multirow{2}*{BLEU} & \multicolumn{2}{c}{$\textrm{SLC}_{p}$} \\
 \cmidrule{3-4} \cmidrule{6-7} \cmidrule{9-10} \cmidrule{12-13}
 & & $0.4$ &$0.2$ & & $0.4$ &$0.2$ & & $0.4$ &$0.2$ & & $0.4$ &$0.2$ \\
\midrule
Transformer & $41.70$ & $74.14 $ &41.89 & $39.92$ & $71.46 $ & $42.19$ & $15.46 $ & $80.33$ & $41.80$ &$15.31 $ & $38.24 $ &  $13.82$ \\
Baseline &$32.61$ & $75.09$ & $41.03$ & $ 31.32 $ & $70.65$ & $41.19$ & $13.60$ & $82.64 $ & $54.37$ & $14.12 $ & $57.76$ & $24.09$ \\
\midrule
VideoDubber (Target) &$41.67$ & $80.30 $ &$52.87$ & $39.10$ &$78.52$ &$48.82 $ & $14.66 $& $91.43$& $68.82$ & $14.32$& $72.19 $ & $60.05$  \\
\midrule
VideoDubber (Source) & $39.60$ & $\textbf{79.39} $& $\textbf{45.31}$ & $36.64$ & $\textbf{75.76} $ & $\textbf{44.83}$ & $13.77 $ & $\textbf{90.10}$ & $ \textbf{57.03} $&$13.95 $ & $\textbf{74.55} $ & $\textbf{41.96}$  \\

\bottomrule
\end{tabular}
\caption{BLEU scores and $SLC_p$ score ($\uparrow$) on CVSS Es-En, CVSS De-En, CVSS Zh-En, MuST-C En-Zh. Here, we set $p = 0.4 $ and $ p = 0.2$ respectively. We compare our method with Transformer~\cite{Transformer2017} (without any length control strategy) and baseline~\cite{federico2020stst2vd} (with text length control).  And ``VideoDubber (Source)'' and ``VideoDubber (Target)'' represents controlling the translation with the length
of the source speech and golden target speech respectively.}
\label{tab1}
\end{table*}

\section{Real-World Video Dubbing Test Set}
Considering the scarcity of real-world video dubbing dataset~(i.e., motion pictures with golden cross-lingual source and target speech), we construct a test set collected from dubbed films to provide comprehensive evaluations of video dubbing systems. 

Specifically, we select nine popular films translated from English to Chinese, which are of high manual translation and dubbing quality, and contain rich genres such as love, action, scientific fiction, etc. We cut 42 conversation clips from them with the following criteria: 1) The clip duration is around 1 $\sim$ 3 minutes. 2) More than 10 sentences are involved in each clip, which contains both long and short sentences. 3) The face of speaker is visible mostly during his or her talks, especially visible lips at the end of speech.

After that, we extract audios from video clips and remove background sounds to get clean speech. Then, we perform speech recognition and manual correction to obtain text transcripts in source and target languages. In addition, we split clips into sentences based on semantics and speakers. We discard silence frames between sentences to make sure there is no more than 0.5s of silence at the beginning and end of each split.

Finally, we obtain the test dataset with a total duration of 1 hour, including the original movie, source speech with transcripts, and human-dubbed speech with transcripts. It contains 892 sentences for sentence-level evaluation\footnote{The test set can be built following \url{https://speechresearch.github.io/videodubbing/}}. 



\section{Experimental Setup}

\subsection{Datasets}
\paragraph{Training Data}
We train VideoDubber on four language directions: Chinese $\rightarrow$ English~(Zh-En), English $\rightarrow$ Chinese~(En-Zh), German $\rightarrow$ English~(De-En), Spanish $\rightarrow$ English~(Es-En).
Since the absence of real-world video dubbing dataset, we train and test VideoDubber in the speech translation dataset. For language direction from Others $\rightarrow$ English,
we use public speech-to-speech translation dataset CVSS~\cite{jia2022cvss}, which contains multilingual-to-English speech-to-speech translation corpora derived from the CoVoST2 dataset~\cite{wang2021covost2}.
For the direction from English to Chinese, we use the En-Zh subset of MuST-C~\cite{cattoni2021mustc}, an English $\rightarrow$ Others speech translation corpus built from English TED Talks. Since the Must-C dataset does not have corresponding speech in the target language, we use a well trained Chinese TTS model, FastSpeech 2~\cite{ren2021fs2}, to generate Chinese speech from the translated text.

\paragraph{Evaluation Data}
We consider two datasets in evaluation.
One is the standard test set from the speech translation dataset that follows the settings of previous works, and the
other is the constructed real-world test set for the video dubbing scenario, to test the overall performance in the real-world setting.

\subsection{Evaluation Metrics}
We adopt both 
subjective and objective evaluation metrics to evaluate the compared video dubbing systems.
\paragraph{Objective Metrics} We use BLEU and speech length compliant to evaluate the semantic quality of the translated text and the length control ability of the model.
\begin{itemize}
    \item \textbf{BLEU.} BLEU~\cite{BLEU2002} score is a widely used automatic evaluation metric of machine translation task. We report the tokenized BLEU score to keep consistent with previous works.
    \item \textbf{Speech Length Compliant.} To evaluate the speech length control performance, inspired by~\citet{lakew2022isometricmt}, we design the \textbf{S}peech \textbf{L}ength \textbf{C}ompliant ($\textrm{SLC}_p$) score.
    $\textrm{SLC}_p$ refers to the percentage of sentences whose $ratio \in \left[ 1-p, 1+p \right ]$, where the ratio is defined as:
    \begin{equation}
        ratio = \frac{\sum_{i=0}^{T_{y^{'}}} d_i}{\sum_{j=0}^{T_{x}} d_j},
    \end{equation}
    where $d_i$, $d_j$ represent the duration of token $i$, $j$ from source speech and translated speech respectively. $\textrm{SLC}_p$ refers to the percentage of sentences that $ratio \in \left[ 1-p, 1+p \right ]$. ~\citet{federico2020evapa} set $p=0.4$ to evaluate the fluency of synthesized audio, while we find that $p=0.2$ is a more reasonable interval for users from the user study (i.e., speaking rate speedup by $1.4x$ or slowdown by $0.6x$ will cause large inconsistency that affects user watching experience). Therefore, we use both $p=0.2$ 
    and $p=0.4$ 
    and compare their corresponding $\textrm{SLC}_p$ in the experiments respectively.
\end{itemize}
\paragraph{Subjective Metrics}
In addition to objective metrics, we also introduce some subjective metrics to evaluate VideoDubber on the real-world video dubbing scenario.
\begin{itemize}
    \item \textbf{Translation quality.} Translation quality measures the quality of translated text of the dubbed video through human evaluation. It should be mentioned that judges will rate the translation quality according to the original video, which is more suitable for video dubbing scenario.
    \item \textbf{Isochronism.} Isochronism score measures whether the original video matches the translated speech in video dubbing scenario.
    \item \textbf{Naturalness.} Naturalness score measures the overall performance of the video with dubbed speech. The judges will consider both  the translation quality and the
    isochronicity.
\end{itemize}

\subsection{Model Configurations}
We follow the Transformer~\cite{Transformer2017} based encoder-decoder framework as the backbone of our translation model. For all settings, we set the hidden dimension as $512$ for the model, $2048$ for the feed-forward layers.
For both encoder and decoder, we use $6$ layers and $8$ heads for the multi-head attention.
The 
duration-aware positional embeddings are added only at the input to the decoder stacks, and the encoder architecture is consistent with the original Transformer.

The duration predictor consists of a two-layer
convolutional network
with ReLU activation, each followed by the layer normalization and the dropout layer. The extra linear layer outputs a scalar, which is the predicted token duration.
Note that this module is jointly trained with the Transformer model to predict the duration of each token with the mean square error (MSE) loss. We predict the duration in
the logarithmic domain, which makes them easier to train. 
It should be mentioned that duration predictor is only used in inference stage, since the ground truth token duration can be directly used in training.

In TTS stage, we employ a pre-trained zero-shot TTS model,
AdaSpeech 4~\cite{ada4}. Given a reference speech, it can synthesize voice for unseen speakers. Here, we regard source speech as reference, generating speech with duration adjustment.  

\subsection{Baseline}
We compare VideoDubber with well applied baseline~\cite{federico2020stst2vd}, which control the length of translated text in NMT by using the number of words. According to the target/source token number ratio, it splits the sentences from training dataset into three groups (short, normal, long). At training phrase, it pretends the corresponding length ratio token to source sequence. At inference phrase, the desired length token (i.e. normal) is pretended to source sequence to control the length of translated text.

\begin{table*}[h]
\small
\centering
\begin{tabular}{l|c|c|c|c|c|c|c|c|c|c|c|c}
\toprule
\multirow{3}*{Settings} &\multicolumn{3}{c|}{Es-En} & \multicolumn{3}{c|}{De-En} & \multicolumn{3}{c|}{Zh-En} &  \multicolumn{3}{c}{En-Zh} \\
\cmidrule{2-13}
 & \multirow{2}*{BLEU} & \multicolumn{2}{c|}{$\textrm{SLC}_{p}$} & \multirow{2}*{BLEU} & \multicolumn{2}{c|}{$\textrm{SLC}_{p}$} & \multirow{2}*{BLEU} & \multicolumn{2}{c|}{$\textrm{SLC}_{p}$}  & \multirow{2}*{BLEU} & \multicolumn{2}{c}{$\textrm{SLC}_{p}$} \\
 \cmidrule{3-4} \cmidrule{6-7} \cmidrule{9-10} \cmidrule{12-13}
 \cmidrule{3-4} \cmidrule{6-7} \cmidrule{9-10} \cmidrule{12-13}
 & & $0.4$ &$0.2$ & & $0.4$ &$0.2$ & & $0.4$ &$0.2$ & & $0.4$ &$0.2$ \\
\midrule

 VideoDubber  & $39.60$ & $\textbf{79.39} $& $\textbf{45.31}$ & $36.64$ & $\textbf{75.76} $ & $\textbf{44.83}$ &$13.77 $ & $\textbf{90.10}$ & $ \textbf{57.03} $&$13.95 $ & $\textbf{74.55} $ & $\textbf{41.96}$  \\
 \hspace{1em} $w/o$ abs PE  & $41.58$ & $75.70 $& $42.88$ & $39.92$ & $71.46 $ &$42.19$ & $15.46 $ & $80.33$ &$41.80$ & $15.31 $ & $42.88$ &$28.43$ \\
 \hspace{1em} $w/o$ rel PE  & $40.53 $ & $ 74.57 $& $42.30$ & $34.83 $ & $ 71.60$ & $39.46$ & $14.73 $ & $82.56 $& $52.34$ &$15.21 $ & $ 54.55$ & $30.50$ \\
 \hspace{1em} $w/o$ original PE  & $39.62 $ & $73.98  $& $41.55 $ & $37.68 $ & $73.46 $ & $ 43.42 $ & $ 13.02 $ & $ 82.38 $& $52.29 $ &$ 14.05 $ & $ 53.95 $ & $29.65 $ \\

\bottomrule
\end{tabular}
\caption{Ablation study on CVSS Es-En, CVSS De-En, CVSS Zh-En, MuST-C En-Zh. ``abs PE'' indicates absolute positional embedding and ``rel PE'' indicates relative positional embedding. Note that unless otherwise stated, ``VideoDubber'' refers to our proposed method which controls the translation with the length of source speech.}
\label{tab2}
\end{table*}

\begin{table*}[h] \small
\centering
\begin{tabular}{l|ccc|ccc}
\toprule
\multirow{2}*{Settings} &\multicolumn{3}{c|}{Objective} & \multicolumn{3}{c}{Subjective}\\
\cmidrule{2-7}
& BLEU & $\textrm{SLC}_{0.4}$  & $\textrm{SLC}_{0.2}$  & Translation Quality & Isochronous & Naturalness \\

\midrule
Transformer~\cite{Transformer2017} &7.21 &41.75\%& 27.25\%& 3.45 &2.97 & 3.00 \\
Baseline~\cite{federico2020stst2vd} &6.56 & 52.17\%&38.16\% &3.64 &3.30 &3.40 \\
\midrule
VideoDubber &\textbf{7.31} &\textbf{79.17\%}& \textbf{66.67\%} & \textbf{3.98} &\textbf{3.41} &\textbf{3.60} \\
\bottomrule
\end{tabular}
\caption{Objective and subjective evaluation on real-world video dubbing test set. We compare our method with Transformer~\cite{Transformer2017} (without any length control strategy) and baseline~\cite{federico2020stst2vd} (with text length control).}
\label{tab3}
\end{table*}

\section{Results}
\subsection{Automatic Evaluation} 
We show the machine translation quality and the length control performance of related models on four language directions in Table~\ref{tab1}.
As discussed in Section~\ref{sec:discuss}, we also list the results when controlling the translation with the length of the golden target speech, to show the upper-bound performance of our model.
When considering the length control ability measured by $\textrm{SLC}_{p}$, VideoDubber consistently outperforms considered baselines with a large margin, illustrating that the proposed speech-aware length control achieves better speech duration isochronism than controlling the numbers of words/characters~\citep{federico2020stst2vd}.
Especially, on the En-Zh direction, we find that VideoDubber brings significant improvements over the token number controlling baseline. The reason may be the large inconsistency between the token number and speech duration in Chinese, i.e., the token in Chinese may contain several Chinese characters instead of only a sub-word in English. As a result, even small mismatch on the token number may cause large difference on the speech duration in Chinese. VideoDubber alleviates this problem by directly controlling the speech length.

Regarding the BLEU scores, although with slightly drop compared with the vanilla Transformer model, VideoDubber outperforms the length control baseline, especially in Es-En and De-En. The baseline method performs coarse length control, with the same length control embedding for all sentences. It may lead to unstable translation ability in some datasets.
In addition, it is worth noting that the source and target sentences in the test set do not have the same speech length. It is unfair to judge the translation quality with BLEU scores. Therefore we propose a new real-world test set and conduct fair comparisons on it, as illustrated in Section~\ref{sec:exp-real}.

\subsection{Ablation Study}
To verify the effectiveness of the duration-aware positional embedding, we conduct ablation studies on three kinds of PEs on four language directions, as shown in Table~\ref{tab2}. 
We can find that the absolute and relative duration PE are both crucial to achieve better speech-aware length control results. 


\subsection{Results on Real-world Test Set}
\label{sec:exp-real}
To compare the performance of related methods on the real-world video dubbing scenario,
we conduct experiments on the real-world test set constructed by us. Results are shown in Table~\ref{tab3}.

With objective evaluation, VideoDubber achieves the highest BLEU and $\textrm{SLC}_{p}$ score compared with Transformer and the token number control baseline. It proves that in the real test set when the speech isochronous is considered, the proposed NMT model with speech-aware length control can achieve better isochronous control ability as well as the translation quality. 

Also we conduct subjective evaluation to evaluate the translation quality, the synchronization with the original film footage, and the overall quality of the synthesized speech.
We hired 8 judges who are good at both Chinese and English, asking them to rate the samples generated from different method on the 5 scale. 
From Table~\ref{tab3}, aligned with the objective evaluation, we observe that VideoDubber achieves the highest translation quality and the isochronicity of speech.
Moreover,
our method achieves significantly better performance regarding the Naturalness score, which reflects the overall quality of the automatic dubbed video. Demo samples are available at \url{https://speechresearch.github.io/videodubbing/}.

\section{Conclusion}
In this paper, we propose VideoDubber, a machine translation model with speech-aware length control for the task of video dubbing. To ensure the translated speech to be well aligned with the original video, we 
directly consider
the speech duration of each token in translation. Firstly, we guide the prediction of each word with the duration information, by representing it as two kinds of positional embeddings. Secondly, we introduce a special pause word \texttt{[P]}, which is inserted between each word, in order to control the speech length more smoothly by considering the prosody. Thirdly, we construct a real-world test set collected from dubbed films to provide more actual evaluations of video dubbing systems. Experimental results demonstrate 
that VideoDubber achieves better 
translation quality and isochronous control ability
than related works. For future work, we plan to construct
a training dataset with the same speech length in the source
and target utterances by knowledge distillation, where the
distilled target sentences are controlled by our NMT model
to have the same speech length with the source sentence.

\bigskip
\pagebreak
\bibliography{aaai23}

\begin{thebibliography}{27}
\providecommand{\natexlab}[1]{#1}

\bibitem[{Cattoni et~al.(2021)Cattoni, {Di Gangi}, Bentivogli, Negri, and
  Turchi}]{cattoni2021mustc}
Cattoni, R.; {Di Gangi}, M.~A.; Bentivogli, L.; Negri, M.; and Turchi, M. 2021.
\newblock MuST-C: A multilingual corpus for end-to-end speech translation.
\newblock \emph{Computer Speech \& Language}, 66: 101155.

\bibitem[{Effendi et~al.(2022)Effendi, Virkar, Barra-Chicote, and
  Federico}]{effendi2022durationtts}
Effendi, J.; Virkar, Y.; Barra-Chicote, R.; and Federico, M. 2022.
\newblock Duration Modeling of Neural TTS for Automatic Dubbing.
\newblock In \emph{ICASSP 2022-2022 IEEE International Conference on Acoustics,
  Speech and Signal Processing (ICASSP)}, 8037--8041. IEEE.

\bibitem[{Federico et~al.(2020{\natexlab{a}})Federico, Enyedi, Barra-Chicote,
  Giri, Isik, Krishnaswamy, and Sawaf}]{federico2020stst2vd}
Federico, M.; Enyedi, R.; Barra-Chicote, R.; Giri, R.; Isik, U.; Krishnaswamy,
  A.; and Sawaf, H. 2020{\natexlab{a}}.
\newblock From Speech-to-Speech Translation to Automatic Dubbing.
\newblock In \emph{Proceedings of the 17th International Conference on Spoken
  Language Translation}, 257--264. Online: Association for Computational
  Linguistics.

\bibitem[{Federico et~al.(2020{\natexlab{b}})Federico, Virkar, Enyedi, and
  Barra-Chicote}]{federico2020evapa}
Federico, M.; Virkar, Y.; Enyedi, R.; and Barra-Chicote, R. 2020{\natexlab{b}}.
\newblock Evaluating and Optimizing Prosodic Alignment for Automatic Dubbing.
\newblock In \emph{INTERSPEECH}, 1481--1485.

\bibitem[{Jia et~al.(2022)Jia, Ramanovich, Wang, and Zen}]{jia2022cvss}
Jia, Y.; Ramanovich, M.~T.; Wang, Q.; and Zen, H. 2022.
\newblock {CVSS} Corpus and Massively Multilingual Speech-to-Speech
  Translation.
\newblock \emph{CoRR}, abs/2201.03713.

\bibitem[{Kruspe(2015)}]{kruspe2015speechdata}
Kruspe, A.~M. 2015.
\newblock Training Phoneme Models for Singing with "Songified" Speech Data.
\newblock In \emph{{ISMIR}}, 336--342.

\bibitem[{Lakew et~al.(2021)Lakew, Federico, Wang, Hoang, Virkar,
  Barra-Chicote, and Enyedi}]{lakew2021mtverb}
Lakew, S.~M.; Federico, M.; Wang, Y.; Hoang, C.; Virkar, Y.; Barra-Chicote, R.;
  and Enyedi, R. 2021.
\newblock Machine translation verbosity control for automatic dubbing.
\newblock In \emph{ICASSP 2021-2021 IEEE International Conference on Acoustics,
  Speech and Signal Processing (ICASSP)}, 7538--7542. IEEE.

\bibitem[{Lakew, Gangi, and Federico(2019)}]{lakew2019controlmt}
Lakew, S.~M.; Gangi, M. A.~D.; and Federico, M. 2019.
\newblock Controlling the Output Length of Neural Machine Translation.
\newblock In \emph{{IWSLT}}. Association for Computational Linguistics.

\bibitem[{Lakew et~al.(2022)Lakew, Virkar, Mathur, and
  Federico}]{lakew2022isometricmt}
Lakew, S.~M.; Virkar, Y.; Mathur, P.; and Federico, M. 2022.
\newblock Isometric mt: Neural machine translation for automatic dubbing.
\newblock In \emph{ICASSP 2022-2022 IEEE International Conference on Acoustics,
  Speech and Signal Processing (ICASSP)}, 6242--6246. IEEE.

\bibitem[{McAuliffe et~al.(2017)McAuliffe, Socolof, Mihuc, Wagner, and
  Sonderegger}]{mcauliffe2017mfa}
McAuliffe, M.; Socolof, M.; Mihuc, S.; Wagner, M.; and Sonderegger, M. 2017.
\newblock Montreal Forced Aligner: Trainable Text-Speech Alignment Using Kaldi.
\newblock In \emph{{INTERSPEECH}}, 498--502. {ISCA}.

\bibitem[{{\"{O}}ktem, Farr{\'{u}}s, and Bonafonte(2019)}]{oktem2019pa4md}
{\"{O}}ktem, A.; Farr{\'{u}}s, M.; and Bonafonte, A. 2019.
\newblock Prosodic Phrase Alignment for Machine Dubbing.
\newblock In Kubin, G.; and Kacic, Z., eds., \emph{Interspeech 2019, 20th
  Annual Conference of the International Speech Communication Association,
  Graz, Austria, 15-19 September 2019}, 4215--4219. {ISCA}.

\bibitem[{Papineni et~al.(2002)Papineni, Roukos, Ward, and Zhu}]{BLEU2002}
Papineni, K.; Roukos, S.; Ward, T.; and Zhu, W. 2002.
\newblock Bleu: a Method for Automatic Evaluation of Machine Translation.
\newblock In \emph{{ACL}}, 311--318. {ACL}.

\bibitem[{Ren et~al.(2021)Ren, Hu, Tan, Qin, Zhao, Zhao, and Liu}]{ren2021fs2}
Ren, Y.; Hu, C.; Tan, X.; Qin, T.; Zhao, S.; Zhao, Z.; and Liu, T. 2021.
\newblock FastSpeech 2: Fast and High-Quality End-to-End Text to Speech.
\newblock In \emph{{ICLR}}. OpenReview.net.

\bibitem[{Sennrich, Haddow, and Birch(2015)}]{sennrich2015neural}
Sennrich, R.; Haddow, B.; and Birch, A. 2015.
\newblock Neural machine translation of rare words with subword units.
\newblock \emph{arXiv preprint arXiv:1508.07909}.

\bibitem[{Sharma et~al.(2021)Sharma, Virkar, Federico, Barra-Chicote, and
  Enyedi}]{sharma2021intratts}
Sharma, M.; Virkar, Y.; Federico, M.; Barra-Chicote, R.; and Enyedi, R. 2021.
\newblock Intra-Sentential Speaking Rate Control in Neural Text-To-Speech for
  Automatic Dubbing.
\newblock In \emph{Interspeech}, 3151--3155.

\bibitem[{Takase and Okazaki(2019)}]{takase2019pecontrol}
Takase, S.; and Okazaki, N. 2019.
\newblock Positional Encoding to Control Output Sequence Length.
\newblock In \emph{{NAACL-HLT} {(1)}}, 3999--4004. Association for
  Computational Linguistics.

\bibitem[{Tam et~al.(2022)Tam, Lakew, Virkar, Mathur, and
  Federico}]{derek2022isochronyMT}
Tam, D.; Lakew, S.~M.; Virkar, Y.; Mathur, P.; and Federico, M. 2022.
\newblock Isochrony-Aware Neural Machine Translation for Automatic Dubbing.
\newblock In Ko, H.; and Hansen, J. H.~L., eds., \emph{Interspeech 2022, 23rd
  Annual Conference of the International Speech Communication Association,
  Incheon, Korea, 18-22 September 2022}, 1776--1780. {ISCA}.

\bibitem[{Tan et~al.(2022)Tan, Chen, Liu, Cong, Zhang, Liu, Wang, Leng, Yi, He
  et~al.}]{tan2022naturalspeech}
Tan, X.; Chen, J.; Liu, H.; Cong, J.; Zhang, C.; Liu, Y.; Wang, X.; Leng, Y.;
  Yi, Y.; He, L.; et~al. 2022.
\newblock NaturalSpeech: End-to-End Text to Speech Synthesis with Human-Level
  Quality.
\newblock \emph{arXiv preprint arXiv:2205.04421}.

\bibitem[{Tan et~al.(2021)Tan, Qin, Soong, and Liu}]{tan2021survey}
Tan, X.; Qin, T.; Soong, F.; and Liu, T.-Y. 2021.
\newblock A survey on neural speech synthesis.
\newblock \emph{arXiv preprint arXiv:2106.15561}.

\bibitem[{Vaswani et~al.(2017{\natexlab{a}})Vaswani, Shazeer, Parmar,
  Uszkoreit, Jones, Gomez, Kaiser, and
  Polosukhin}]{DBLP:conf/nips/VaswaniSPUJGKP17}
Vaswani, A.; Shazeer, N.; Parmar, N.; Uszkoreit, J.; Jones, L.; Gomez, A.~N.;
  Kaiser, L.; and Polosukhin, I. 2017{\natexlab{a}}.
\newblock Attention is All you Need.
\newblock In \emph{{NIPS}}, 5998--6008.

\bibitem[{Vaswani et~al.(2017{\natexlab{b}})Vaswani, Shazeer, Parmar,
  Uszkoreit, Jones, Gomez, Kaiser, and Polosukhin}]{Transformer2017}
Vaswani, A.; Shazeer, N.; Parmar, N.; Uszkoreit, J.; Jones, L.; Gomez, A.~N.;
  Kaiser, L.; and Polosukhin, I. 2017{\natexlab{b}}.
\newblock Attention is All you Need.
\newblock In Guyon, I.; von Luxburg, U.; Bengio, S.; Wallach, H.~M.; Fergus,
  R.; Vishwanathan, S. V.~N.; and Garnett, R., eds., \emph{Advances in Neural
  Information Processing Systems 30: Annual Conference on Neural Information
  Processing Systems 2017, December 4-9, 2017, Long Beach, CA, {USA}},
  5998--6008.

\bibitem[{Virkar et~al.(2021)Virkar, Federico, Enyedi, and
  Barra-Chicote}]{virkar2021improvepa}
Virkar, Y.; Federico, M.; Enyedi, R.; and Barra-Chicote, R. 2021.
\newblock Improvements to prosodic alignment for automatic dubbing.
\newblock In \emph{ICASSP 2021-2021 IEEE International Conference on Acoustics,
  Speech and Signal Processing (ICASSP)}, 7543--7574. IEEE.

\bibitem[{Virkar et~al.(2022)Virkar, Federico, Enyedi, and
  Barra-Chicote}]{virkar2022offscreenpa}
Virkar, Y.; Federico, M.; Enyedi, R.; and Barra-Chicote, R. 2022.
\newblock Prosodic Alignment for off-screen automatic dubbing.
\newblock \emph{arXiv preprint arXiv:2204.02530}.

\bibitem[{Wahlster(2013)}]{wahlster2013verbmobil}
Wahlster, W. 2013.
\newblock \emph{Verbmobil: foundations of speech-to-speech translation}.
\newblock Springer Science \& Business Media.

\bibitem[{Wang et~al.(2021)Wang, Wu, Gu, and Pino}]{wang2021covost2}
Wang, C.; Wu, A.; Gu, J.; and Pino, J. 2021.
\newblock CoVoST 2 and Massively Multilingual Speech Translation.
\newblock In Hermansky, H.; Cernock{\'{y}}, H.; Burget, L.; Lamel, L.;
  Scharenborg, O.; and Motl{\'{\i}}cek, P., eds., \emph{Interspeech 2021, 22nd
  Annual Conference of the International Speech Communication Association,
  Brno, Czechia, 30 August - 3 September 2021}, 2247--2251. {ISCA}.

\bibitem[{Wu et~al.(2022)Wu, Tan, Li, He, Zhao, Song, Qin, and Liu}]{ada4}
Wu, Y.; Tan, X.; Li, B.; He, L.; Zhao, S.; Song, R.; Qin, T.; and Liu, T. 2022.
\newblock AdaSpeech 4: Adaptive Text to Speech in Zero-Shot Scenarios.
\newblock In Ko, H.; and Hansen, J. H.~L., eds., \emph{Interspeech 2022, 23rd
  Annual Conference of the International Speech Communication Association,
  Incheon, Korea, 18-22 September 2022}, 2568--2572. {ISCA}.

\bibitem[{Yu and Deng(2016)}]{yu2016automatic}
Yu, D.; and Deng, L. 2016.
\newblock \emph{Automatic speech recognition}, volume~1.
\newblock Springer.

\end{thebibliography}
\newpage
\appendix

\end{document}